\pdfoutput=1

\documentclass[11pt]{article}

\usepackage[]{acl}

\usepackage{times}
\usepackage{latexsym}

\usepackage[T1]{fontenc}

\usepackage[utf8]{inputenc}

\usepackage{microtype}

\usepackage{booktabs}
\usepackage{graphics}
\usepackage{multirow}
\usepackage{tablefootnote}
\usepackage{amsmath}
\usepackage{mathtools}
\usepackage{wrapfig}
\usepackage{amsmath,amsfonts}

\newcommand{\cready}[1]{}
\newcommand{\sr}[3]{$#1\!#2\!#3$}   

\newcommand{\sib}{sIB}
\newcommand{\itmlm}{BERT$_{\textrm{IT:MLM}}$}
\newcommand{\itclust}{BERT$_{\textrm{IT:CLUST}}$}
\newcommand{\itmlmclust}{BERT$_{\textrm{IT:MLM+CLUST}}$}
\newcommand{\glovesvm}{SVM$_{\textrm{GloVe}}$}
\usepackage{placeins}
\usepackage{xr}

\makeatletter
\newcommand*{\addFileDependency}[1]{
  \typeout{(#1)}
  \@addtofilelist{#1}
  \IfFileExists{#1}{}{\typeout{No file #1.}}
}
\makeatother

\newcommand*{\myexternaldocument}[1]{%
    \externaldocument{#1}%
    \addFileDependency{#1.tex}%
    \addFileDependency{#1.aux}%
}
\myexternaldocument{9_ACL_supplemental}

\title{Cluster \& Tune: Boost Cold Start Performance in Text Classification}

\author{\bf{Eyal Shnarch\thanks{\ \ These authors contributed equally to this work.}, Ariel Gera\footnotemark[1], Alon Halfon\footnotemark[1], Lena Dankin, Leshem Choshen, Ranit Aharonov,} \\
\bf{Noam Slonim} \\
\\
IBM Research \\
\{eyals, alonhal, lenad, leshem.choshen, noams\}@il.ibm.com,
\\ariel.gera1@ibm.com, ranitah1@gmail.com}

\begin{document}
\maketitle
\begin{abstract}
In real-world scenarios, a text classification task often begins with a \textit{cold start}, when labeled data is scarce. In such cases, the common practice of fine-tuning pre-trained models, such as BERT, for a target classification task, is prone to produce poor performance.  
We suggest a method to boost the performance of such models by adding an intermediate unsupervised classification task, between the pre-training and fine-tuning phases. 
As such an intermediate task, we perform clustering and train the pre-trained model on predicting the cluster labels.
We test this hypothesis on various data sets, and show that this additional classification phase can significantly improve performance, 
mainly for topical classification tasks,
when the number of labeled instances available for fine-tuning is only a couple of dozen to a few hundred. 
\end{abstract}

\section{Introduction} \label{sec:intro}

The standard paradigm for text classification relies on supervised learning, where it is well known that the size and quality of the labeled data strongly impact the performance \citep{Raffel2019ExploringTL}. Hence, developing a text classifier in practice typically requires making the most of a relatively small set of annotated examples. 

The emergence of transformer-based pre-trained language models such as BERT \citep{BERT} has reshaped the NLP landscape, leading to significant advances in the performance of most NLP tasks, text classification included (e.g., \citealp{nogueira2019passage, ein2020corpus}). These models typically rely on pretraining with massive and heterogeneous corpora on a general 
Masked Language Modeling (\textit{MLM}) task, i.e., predicting a word that is masked in the original text. Later on, the obtained model is fine-tuned to the actual task of interest, termed here the \textit{target task}, using the labeled data available for this task. Thus, pretrained models serve as general sentence encoders which can be adapted to 
a variety of target tasks \citep{Lacroix2019NoisyCF, Wang2020ToPO}. 

Our work focuses on a challenging yet common scenario, where unlabeled data is available but labeled data is scarce. In many real-world scenarios, obtaining even a couple of hundred of labeled examples per class is challenging. 
Commonly, a target class has a relatively low prior in the examined data, making it a formidable goal to collect enough positive examples for it \citep{japkowicz2002class}.
Moreover, sometimes data cannot be labeled via crowd-annotation platforms due to its confidentiality (be it for data privacy reasons or for protecting intellectual property) or since the labeling task requires special expertise.
On top of this, often the number of categories to be considered is relatively large, e.g., $50$, thus making even a modest demand of 200 labeled examples per class a task of labeling 10K instances, which is inapplicable in many practical cases (for an extreme example, cf. \citealp[]{partalas2015lshtc}).

In such limited real-world settings, fine-tuning a 
large pretrained model often yields 
far from optimal performance. 
To overcome this, one may take a gradual approach composed of various phases. One possibility is to further pretrain the model with the {\it self-supervised\/} MLM task over unlabeled data taken from the target task domain \citep{Whang2019DomainAT}. Alternatively, one can train the pretrained model using a {\it supervised\/} intermediate task which is different in nature from the target-task, and for which labeled data is more readily available \citep{pruksachatkun2020intermediatetask,wang-etal-2019-tell,phang2018sentence}.
Each of these steps is expected to provide a better starting point 
for the final fine-tuning phase, performed over the scarce labeled data available for the target task, aiming to end up with improved performance. 

Following these lines, here we propose a 
strategy that exploits {\it unsupervised\/} text clustering as the intermediate task towards fine-tuning a pretrained model for text classification. Our work is inspired by the use of clustering to obtain labels in computer vision \citep{Gidaris2018UnsupervisedRL, Kolesnikov2019RevisitingSV}. Specifically, we use an efficient clustering technique, that relies on simple Bag Of Words (BOW) representations, to partition the unlabeled training data into relatively homogeneous clusters of text instances.
Next, we treat these clusters as labeled data for an intermediate text classification task, and train the pre-trained model -- with or without additional MLM pretraining -- with respect to this multi-class problem, prior to the final fine-tuning over the actual target-task labels. Extensive experimental results demonstrate the practical value of this strategy on a variety of benchmark data. We further analyze the results to gain insights as to why and when this approach would be most valuable, and conclude that it is most prominently when the training data available for the target task is relatively small and the classification task is of a topical nature. Finally, we propose future directions. 

We release code for reproducing our method.\footnote{\url{https://github.com/IBM/intermediate-training-using-clustering}}
\section{Intermediate Training using Unsupervised Clustering} \label{sec:inter}

\begin{figure*}[th]
\begin{center}
\includegraphics[width=0.98\textwidth, trim={0.4cm 0.6cm 0.7cm 0.6cm},clip]{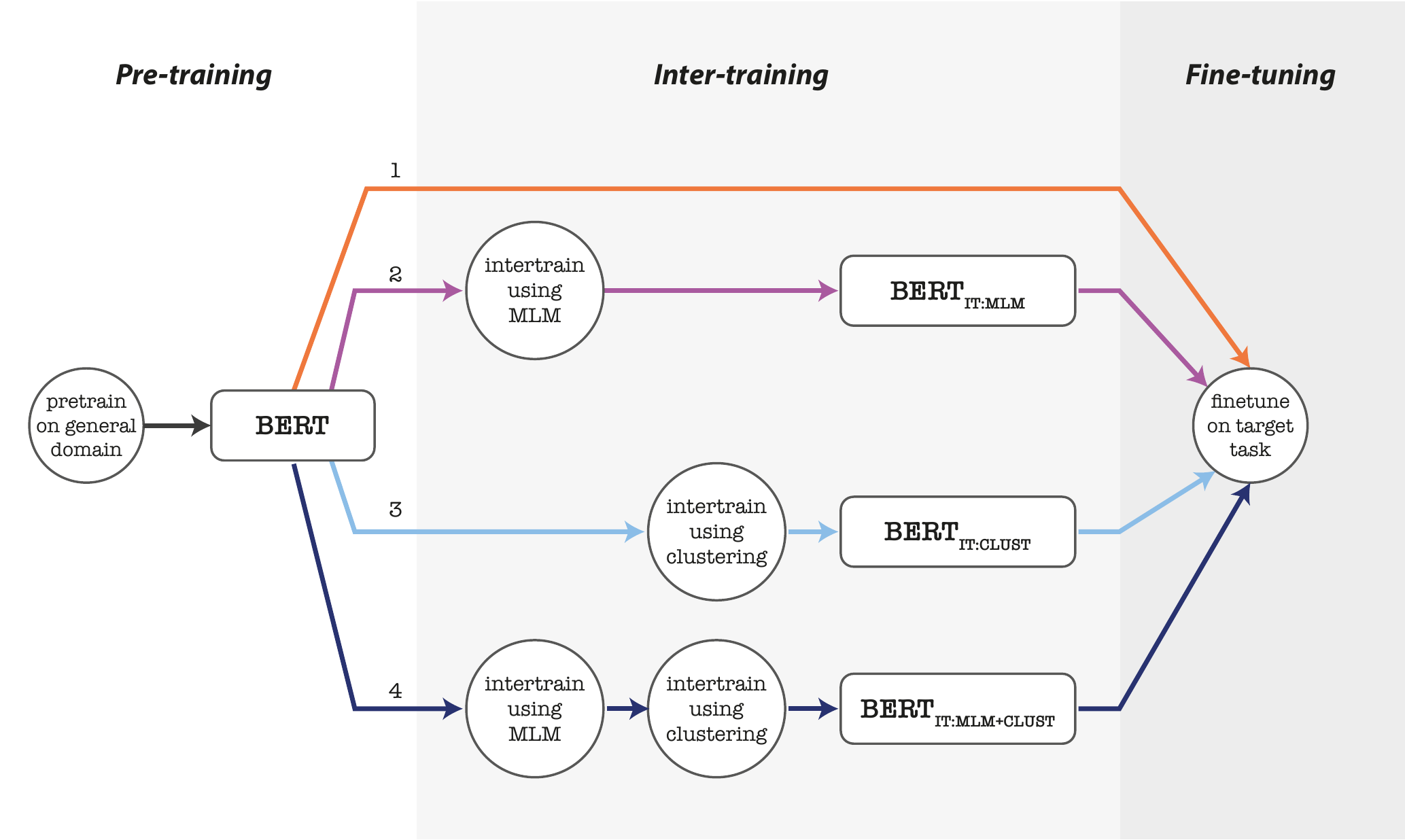}
\caption{Phases of a pre-trained model (BERT in this figure) - circles are training steps which produce models, represented as rectangles.
In the pre-training phase, only general corpora are available. The inter-training phase is exposed to target domain data, but not to its labeled instances. Those are only available at the fine-tuning phase.}
\label{fig:models}
\end{center}
\end{figure*}

A pre-trained model is typically developed in consecutive phases. 
Henceforth, we will refer to BERT as the canonical example of such models.
First, the model is \textit{pretrained} over massive general corpora with the MLM task.\footnote{BERT was originally also pretrained over "next sentence prediction"; however, later works \cite{Yang2019XLNetGA,liu2019roberta} have questioned the contribution of this additional task and focused on MLM.} 
We denote the obtained model simply as {\it BERT\/}. 
Second, BERT is {\it finetuned\/} in a supervised manner 
with the available labeled examples for the target task at hand. This 
standard flow is represented via Path-1 in Fig. \ref{fig:models}. 

An additional phase can be added between these two, referred to next as \textit{intermediate training}, or inter-training in short. In this phase, the model is exposed to the corpus of the target task, or a corpus of the same domain, but still has no access to labeled examples for this task. 

A common example of such an intermediate phase is to continue to intertrain BERT using the self-supervised MLM task over the corpus or the domain of interest, sometimes referred to as further or adaptive pre-training (e.g., \citealp{Gururangan2020DontSP}). 
This flow is represented via Path-2 in Fig. \ref{fig:models}, and the resulting model is denoted \textit{\itmlm{}}, standing for Intermediate Task: MLM.

A key contribution of this paper is to propose a new type of intermediate task, which is designed to be aligned with a text classification target task, and is straightforward to use in practice. The underlying intuition is that inter-training the model over a related text classification task would be more beneficial compared to MLM inter-training, which focuses on different textual entities, namely predicting the identity of a single token. 

Specifically, we suggest {\it unsupervised\/} clustering for generating pseudo-labels for inter-training. As the clustering partition presumably captures information about salient features in the corpus, feeding this information into the model could lead to representations that are better geared to perform the target task.
These pseudo-labels can be viewed as weak labels, but importantly they are not 
tailored nor require a specific design per target task. Instead, we suggest generating pseudo-labels in a way independent of the target classification task.
The respective flow is represented via Path-3 in Fig. \ref{fig:models}.
In this flow, we first cluster to partition the training data into $n_c$ clusters. Next, we use the obtained partition as `labeled' data in a text classification task, where the classes are defined via the $n_c$ clusters, and intertrain BERT to predict the cluster label. In line with MLM, inter-training includes a classifier layer on top of BERT, which is discarded before the fine-tuning stage. 
The resulting inter-trained model is denoted \textit{\itclust{}}.  

Finally, Path-4 in Fig. \ref{fig:models} represents a sequential composition of Paths 2 and 3. In this flow, we first intertrain BERT with the MLM task. Next, the obtained model is further intertrained to predict the $n_c$ clusters, as in Path-3. The model resulting from this hybrid approach is denoted \textit{\itmlmclust{}}. 

Importantly, following Path-3 or Path-4 requires no additional labeled data, and involves an {\it a-priori\/} clustering of training instances that naturally gives rise to an alternative or an additional inter-training task. As we show in the following sections, despite its simplicity, this strategy provides a significant boost in performance, 
especially when labeled data for the final fine-tuning is in short supply. 

\section{Experiments} \label{sec:exp}

\subsection{Tasks and Datasets} \label{ssec:datasets}

We evaluate over $6$ topical datasets and $3$ non-topical ones (see Table~\ref{tab:datasets_paper}), which cover a variety of classification tasks and domains: \href{http://groups.di.unipi.it/~gulli/AG_corpus_of_news_articles.html}{Yahoo! Answers} \citep{ds-ag-news:15}, which separates answers and questions to types;
\href{http://groups.di.unipi.it/~gulli/AG_corpus_of_news_articles.html}{DBpedia} (\citealp{ds-ag-news:15}, CC-BY-SA) which differentiates 
entity types by their Wikipedia articles;
\href{http://groups.di.unipi.it/~gulli/AG_corpus_of_news_articles.html}{AG’s News} \citep{ds-ag-news:15} which categorize news articles;
\href{https://www.consumerfinance.gov/data-research/consumer-complaints/}{CFPB}, which classifies consumer complaints;
\href{http://qwone.com/~jason/20Newsgroups/}{20 newsgroups} \citep{Lang95newsgroup}, which classifies 20 Usenet discussion groups; 
\href{https://www.unige.ch/cisa/research/materials-and-online-research/research-material/}{ISEAR} (\citealp{ds-isear:15}, CC BY-NC-SA 3.0), which considers personal reports for emotion;
\href{http://www.dt.fee.unicamp.br/~tiago/smsspamcollection/}{SMS spam} \citep{ds-sms:11}, which identifies spam messages;
\href{http://www.cs.cornell.edu/people/pabo/movie-review-data/}{Polarity} \citep{ds-polarity:05}, which includes sentiment analysis on movie reviews,
and \href{http://www.cs.cornell.edu/people/pabo/movie-review-data/}{Subjectivity} \citep{ds-subjectivity:04}, which categorizes movie snippets as subjective or objective.

A topical dataset splits sentences by a high-level distinction related to what the sentence is about (e.g., sports vs. economics). Non-topical datasets look for finer stylistic distinctions that may depend on the way the sentence is written or on fine details rather than on the central meaning it discusses. It may also separate almost identical sentences; for example, "no" could distinguish between sentences with negative and positive sentiment.

When no split is provided we apply a $70\%/10\%/20\%$ train-dev-test split, respectively.\footnote{The dev set is not being used by any method.}
To reduce the computational cost over the larger datasets (DBpedia, AG's News, Yahoo! Answers and CFPB) we trim the train/test sets of these datasets to $15K/3K$ instances respectively, by randomly sampling from each set.\footnote{We verified that relying on the full dataset provides no significant performance improvements to \itmlm{} and \itclust{}. The results are omitted for brevity.} All runs and all methods use only the trimmed versions.

\subsection{Experimental Setup} \label{ssec:exp_setup}
In our main set of experiments, we compare the performance of fine-tuning BERT-based models over a target task, for different settings of intermediate training.
We consider four BERT-based settings, as described in Section \ref{sec:inter} and in Figure \ref{fig:models}.
Two baselines -- (i) BERT, without intermediate training, and (ii) \itmlm{} intertrained on MLM; 
and two settings that rely on clustering -- (i) \itclust{}, where predicting cluster labels is used for inter-training, and (ii) \itmlmclust{}, which combines the two intermediate tasks. 

\paragraph{Training samples:}
For each setting, the final fine-tuning for the target task is performed, per dataset, for training budgets varying between $64$ and $1024$ labeled examples. 
For each data size $x$, the experiment is repeated $5$ times; each repetition representing a different sampling of $x$ labeled examples from the train set. 
The samplings of training examples are shared between all settings. That is, for a given dataset and train size the final training for all settings is done with respect to the same $5$ samples of labeled examples.

\paragraph{Inter-training:} 
Intermediate training, when done, was performed over the unlabeled train set for each dataset (ignoring instances' labels). We studied two implementations for the clustering task: K-means \citep{lloyd1982least} and sequential Information Bottleneck (\sib{}) which is known to obtain better results in practice \citep{sib:02} and in theory \citep{hartigan}.
Based on initial experiments, and previous insights from works in the computer vision domain \citep{Yan2020ClusterFitIG} we opted for a relatively large number of clusters, and rather than optimizing the number of clusters per dataset, set it to $50$ for all cases.\footnote{Setting the number of clusters to be equal to the number of classes resulted in inferior accuracy. In addition,  one may not know how many classes truly exist in the data, so this parameter is not necessarily known in real-world applications.} 
K-means was run over GloVe \citep{Pennington2014GloveGV} representations following word stemming. We used a publicly available implementation of \sib{}\footnote{\url{https://github.com/IBM/sib}} with 
its default configuration (i.e., $10$ restarts and a maximum of $15$ iterations for every single run).
For \sib{} clustering, we used Bag of Words (BOW) representations on a stemmed text with the 
default vocabulary size (which is defined as the 10K most frequent words in the dataset).
Our results indicate that inter-training with respect to \sib{} clusters consistently led to better results in the final performance on the target task, compared to inter-training with respect to the clusters obtained with K-means (see Section \ref{ssec:kmeans} for details).
We also considered inter-training only on representative examples of clustering results -- filtering a given amount of outlier examples -- but obtained no significant gain (data not shown).

Note that the run time of the clustering algorithms is only a few seconds.
The run time of the fine-tuning step of the inter-training task takes five and a half minutes for the largest train set (15K instances) on a Tesla V100-PCIE-16GB GPU.

\paragraph{BERT hyper-parameters:}
The starting point of all settings is the BERT\textsubscript{BASE} model (110M parameters). BERT inter-training and fine-tuning runs were all performed using the Adam optimizer \citep{Kingma2015AdamAM} with a standard setting consisting of a learning rate of $3\times10^{-5}$, batch size $64$, and maximal sequence length $128$.

In a practical setting with a limited annotations budget one cannot assume that a labeled dev set is available, thus in all settings we did not use the dev set, and fine-tuning was arbitrarily set to be over $10$ epochs, always selecting the last epoch.
For inter-training over the clustering results we used a single epoch, for two reasons. First, loosely speaking, additional training over the clusters may drift the model too far towards learning the partition into clusters, which is an auxiliary task in our context, and not the real target task. Second, from the perspective of a practitioner, single epoch training is preferred since it is the least demanding in terms of run time. 
For \itmlm{} we used $30$ epochs with a replication rate of $5$, and followed the masking strategy from \citet{BERT}.\footnote{In preliminary experiments we found this to be the best configuration for this baseline.}

\paragraph{Computational budget:}
Overall we report the results of $1440$ BERT fine-tuning runs ($4$ experimental settings $\times$ $9$ datasets $\times$ 8 labeling budgets $\times$ 5 repetitions). In addition, we performed 288 inter-training epochs over the full datasets ($9$ datasets $\times$ ($30$ \itmlm{} epochs + 1 \itclust{} epoch + 1 \itmlmclust{} epoch)). In total, this would equate to about 60 hours on a single Tesla V100-PCIE-16GB GPU.

\begin{table}
\small
\centering
\begin{tabular}{@{}lrrcc@{}}
\toprule
               & Train  & Test & \begin{tabular}[c]{@{}l@{}}\# classes\end{tabular}\\ \midrule

Yahoo! answers & 15K & 3K & 10      \\
DBpedia       & 15K & 3K & 14    \\
CFPB           & 15K & 3K & 15  \\
20 newsgroups   & 10.2K  & 7.5K & 20\\
AG's news      & 15K  & 3K & 4 \\
ISEAR          & 5.4K     & 1.5K & 7   \\ \midrule
SMS spam       & 3.9K    & 1.1K & 2\\
Subjectivity   & 7K   & 2K & 2 \\
Polarity       & 7.5K   & 2.1K & 2   \\
 \bottomrule
\end{tabular}
\caption{Dataset details. Topical datasets are at the top. \label{tab:datasets_paper}}
\end{table}

\section{Results} \label{sec:results}

\begin{figure*}[t]
\center
\makebox[\textwidth][c]{
  \includegraphics[width=.33\textwidth]{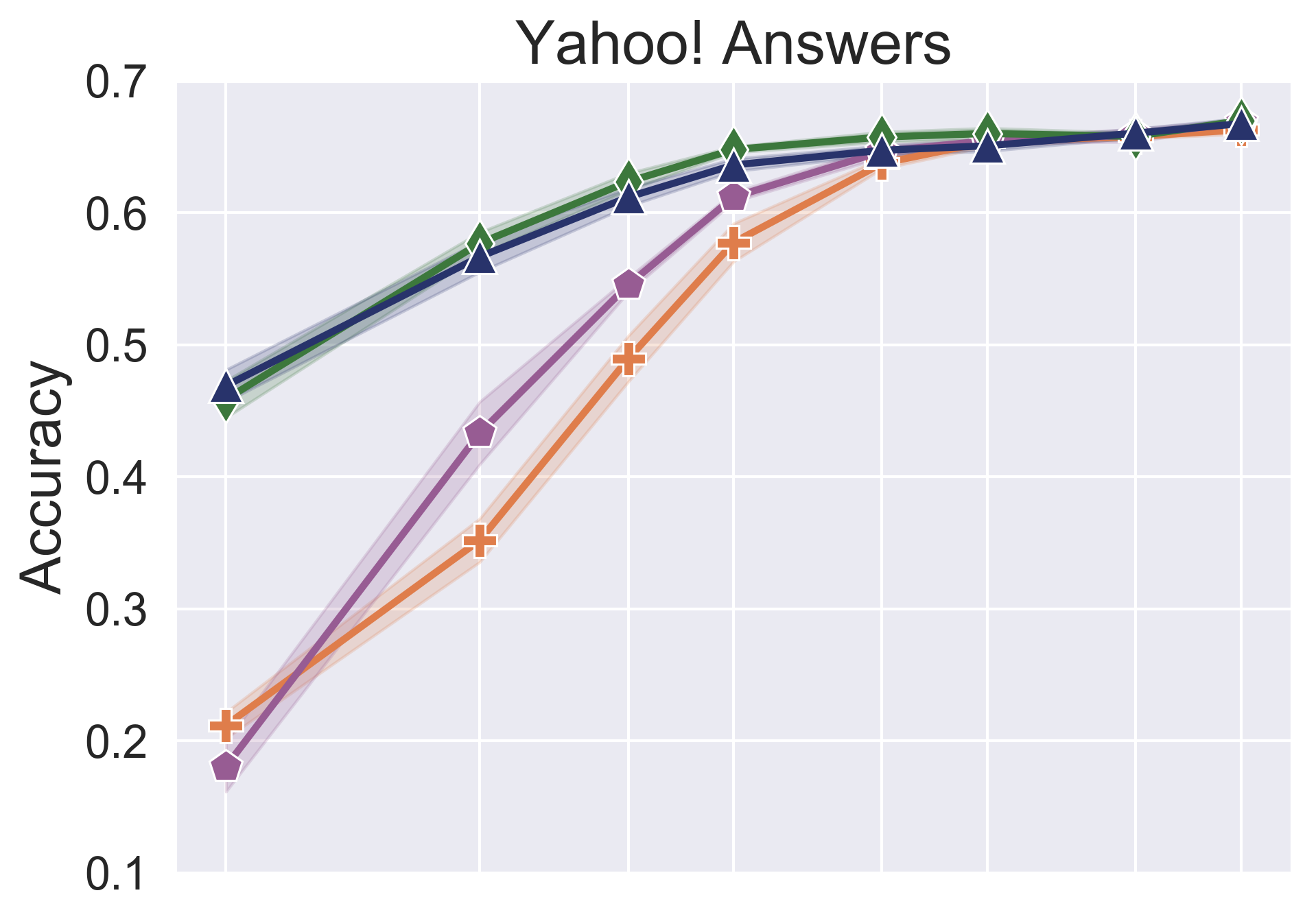}
  \includegraphics[width=.33\textwidth]{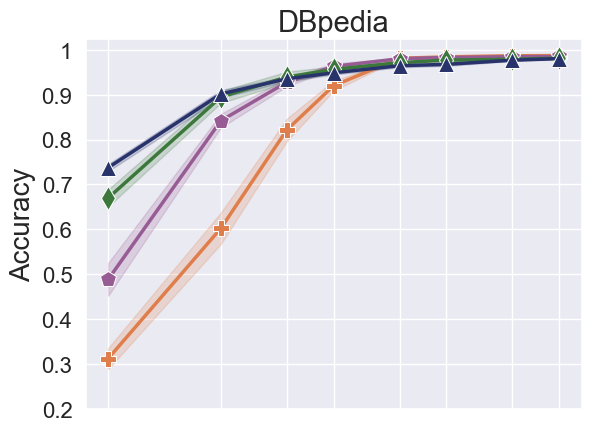}
  \includegraphics[width=.33\textwidth]{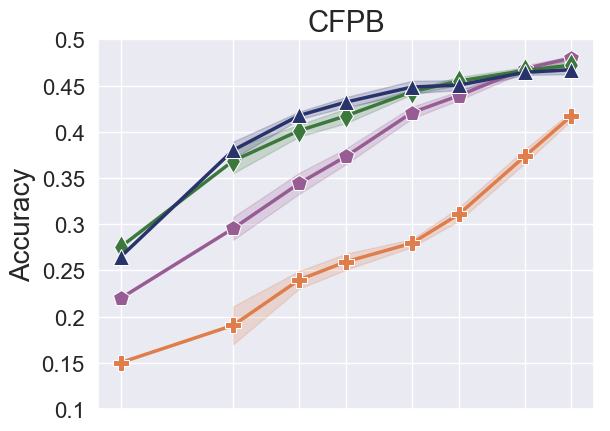}
  }
  
\makebox[\textwidth][c]{
  \includegraphics[width=.33\textwidth]{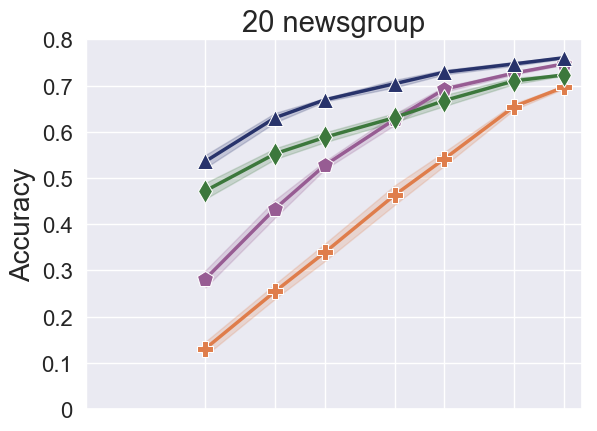}
  \includegraphics[width=.33\textwidth]{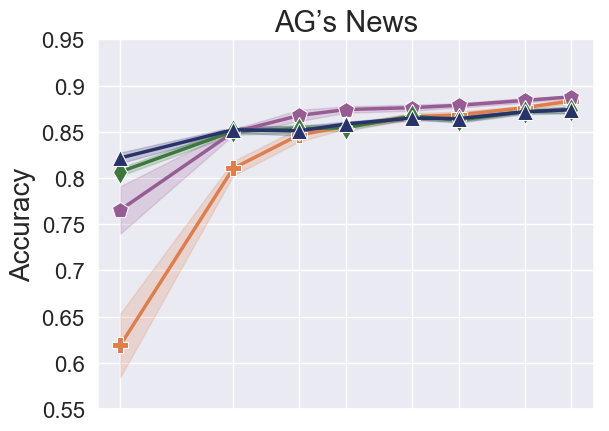}
    \includegraphics[width=.33\textwidth]{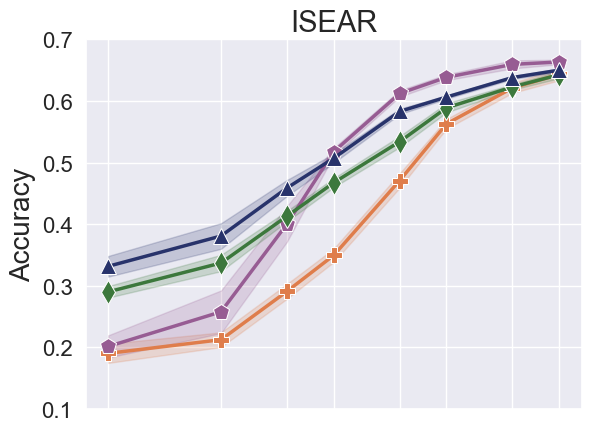}
  }
  
 \makebox[\textwidth][c]{
     \includegraphics[width=.33\textwidth]{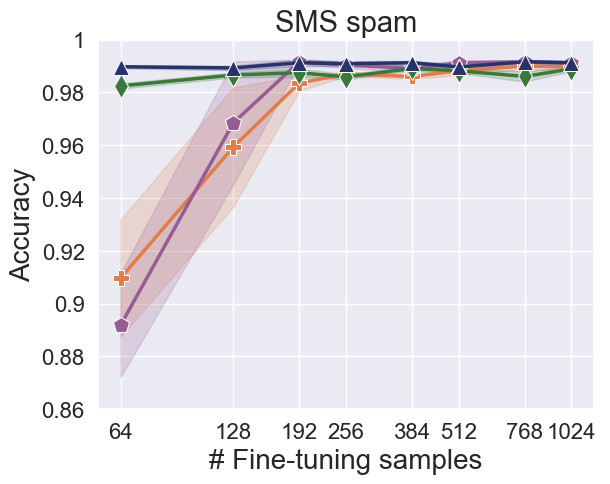}
     \includegraphics[width=.33\textwidth]{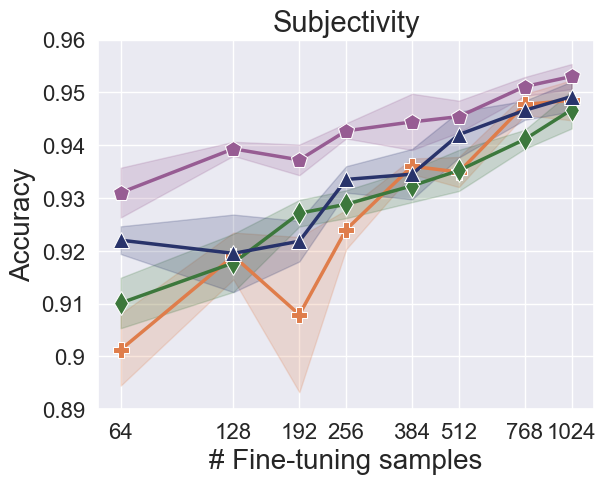}
     \includegraphics[width=.33\textwidth]{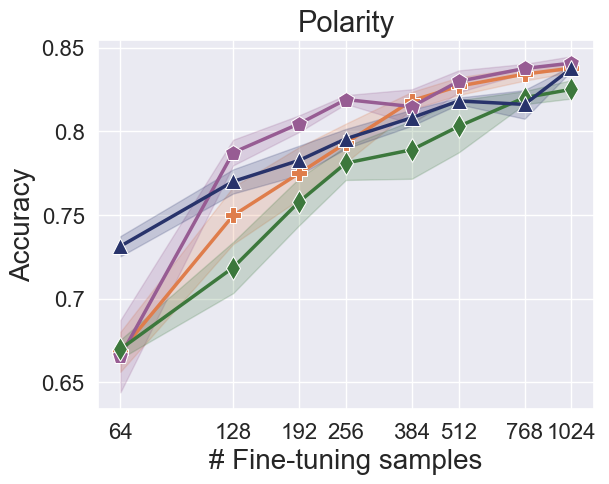}
 }
 
\makebox[\textwidth][c]{
    \includegraphics[width=\textwidth]{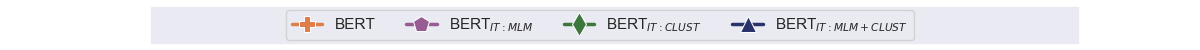}
}
\caption{Classification accuracy ($\pm$SEM, standard error of the mean) on all datasets vs. the number of labeled samples used for fine-tuning (log scale). Each point is the average of 5 repetitions
(for 20 newsgroups and a budget of $64$, all 5 repetitions did not cover all classes and hence this data point is not presented).}

\label{fig:results-plots}
  
\end{figure*}

Table~\ref{tab:focus_compare} depicts the results over all datasets, focusing on the practical use case of a budget of $64$ samples for fine-tuning (128 for 20 newsgroup, see explanation in Fig.~\ref{fig:results-plots}).
As shown in the table, the performance gains of \itclust{} are mainly reflected in the 6 topical datasets. For these datasets, \itclust{} confers a significant benefit in accuracy ($110\%$ accuracy gain, $33\%$ error reduction). 

\begin{table}
\centering
\resizebox{0.5\textwidth}{!}{%
\begin{tabular}{lcccc}
\toprule
\begin{tabular}{@{}c@{}}Dataset\end{tabular} &
\begin{tabular}[c]{@{}c@{}}BERT \\ accuracy \end{tabular} &
\begin{tabular}[c]{@{}c@{}}\itclust{} \\ accuracy \end{tabular} & 
\begin{tabular}{@{}c@{}}Gain\end{tabular} &
\begin{tabular}{@{}c@{}}Error\\reduction\end{tabular}
\\
\midrule
\begin{tabular}{@{}c@{}}Yahoo! Answers\end{tabular} & 21.2 & 45.9 & 117\% & 31\%
\\
DBpedia & 31.2 & 67.0 & 115\% & 52\% \\
CFPB & 15.0 & 27.5 & 83\% & 15\% \\
20 newsgroup & 13.0 & 47.2 & 263\% & 39\% \\
AG’s News & 61.9 & 80.7 & 30\% & 49\%\\
ISEAR & 19.0 & 29.0 & 53\% & 12\%\\
 \hline
 avg. topical & 26.9 & 49.6 & \textbf{110}\% &\textbf{33\%}\\
 \hline
SMS spam & 91.0 & 98.2 & 8\% & 80\%\\
Subjectivity & 90.1 & 91.0 & 1\% & 9\%\\
Polarity & 66.8 & 67.0 & 0\% &1\%\\
\hline
 avg.  non-topical & 82.6 & 85.4 & \textbf{3}\% &\textbf{30\%}\\
\bottomrule
\end{tabular}
}
\caption{\itclust{} outperforms BERT in topical datasets. Comparing 64 samples, the smallest amount for fine-tuning. The accuracy gain and the error reduction (1-accuracy) are relative to BERT's accuracy/error.
\label{tab:focus_compare}}
\end{table}

\begin{table*}
\begin{tabular}{lccccccc}
\toprule
Train size & 64 & 128 & 192 & 256 & 384 & 512 & $>$512\\
\midrule
vs. BERT   & \sr{1}{\times}{10^{-6}} & \sr{1}{\times}{10^{-6}} & \sr{6}{\times}{10^{-7}} & \sr{2}{\times}{10^{-5}} & \sr{2}{\times}{10^{-3}} & \sr{9}{\times}{10^{-3}} & -- \\

vs. \itmlm{}  & \sr{8}{\times}{10^{-5}} & \sr{3}{\times}{10^{-3}} & \sr{4}{\times}{10^{-2}} & -- & -- & -- & -- \\
\bottomrule
\end{tabular}
\caption{Paired t-test p-values (after Bonferroni correction) of classification accuracy for \itclust{} compared to BERT and to \itmlm{} (insignificant results, $p\geq 0.05$, are denoted by --).}
\label{tab:ttest}
\end{table*}

Figure \ref{fig:results-plots} depicts the classification accuracy for the different settings for varying labeling budgets, using \sib{} for clustering-based inter-training.
Over the topical datasets,
\itclust{} and \itmlmclust{}
clearly outperform BERT and \itmlm{} in the small labeled data regime, where the gain is most prominent for the smallest labeled data examined -- when only $64$ labeled examples are available -- and gradually diminishes as more labeled samples are added. 

We performed paired t-tests to compare \itclust{} with BERT and \itmlm{}, pooling together all datasets and repetitions for a given labeling budget. As can be seen in Tab. \ref{tab:ttest}, the performance gain, over all datasets, of \itclust{} over BERT is statistically significant for a budget up to $512$.

\itclust{} is not as successful in the 3 non-topical datasets (cf. Tab.~\ref{tab:focus_compare} and Fig.~\ref{fig:results-plots}). A possible reason for the lack of success of inter-training in these three datasets is that their classification task is different in nature than the tasks in the other six datasets. Identifying spam messages, determining whether a text is subjective or objective, or analyzing the sentiment (polarity) of texts, can be based on stylistic distinctions that may depend on the way the sentence is written rather than on the central topic it discusses. Inter-training over BOW clustering seems to be less beneficial when such considerations are needed. We further analyze this in Section~\ref{ssec:NMI}.  
Nevertheless, it is safe to apply \itclust{} even in these datasets, as results are typically comparable to the baseline algorithms, neither better nor worse.

Both \itmlm{} and \itclust{} expose the model to the target corpus. The performance gains of \itclust{} over \itmlm{} suggest that inter-training on top of the clustering carries an additional benefit. In addition, these inter-training approaches are complementary - as seen in Fig.~\ref{fig:results-plots}, \itmlmclust{} outperforms both \itclust{} and \itmlm{} (at the cost of some added runtime).

Taken together, our results suggest that in topical datasets, 
where labeled data is scarce, the pseudo-labels generated via clustering can be leveraged to provide a better starting point for a pre-trained model towards its fine-tuning for the target task.

\section{Analysis} \label{sec:analysis}

\begin{figure*}

\includegraphics[width=\textwidth]{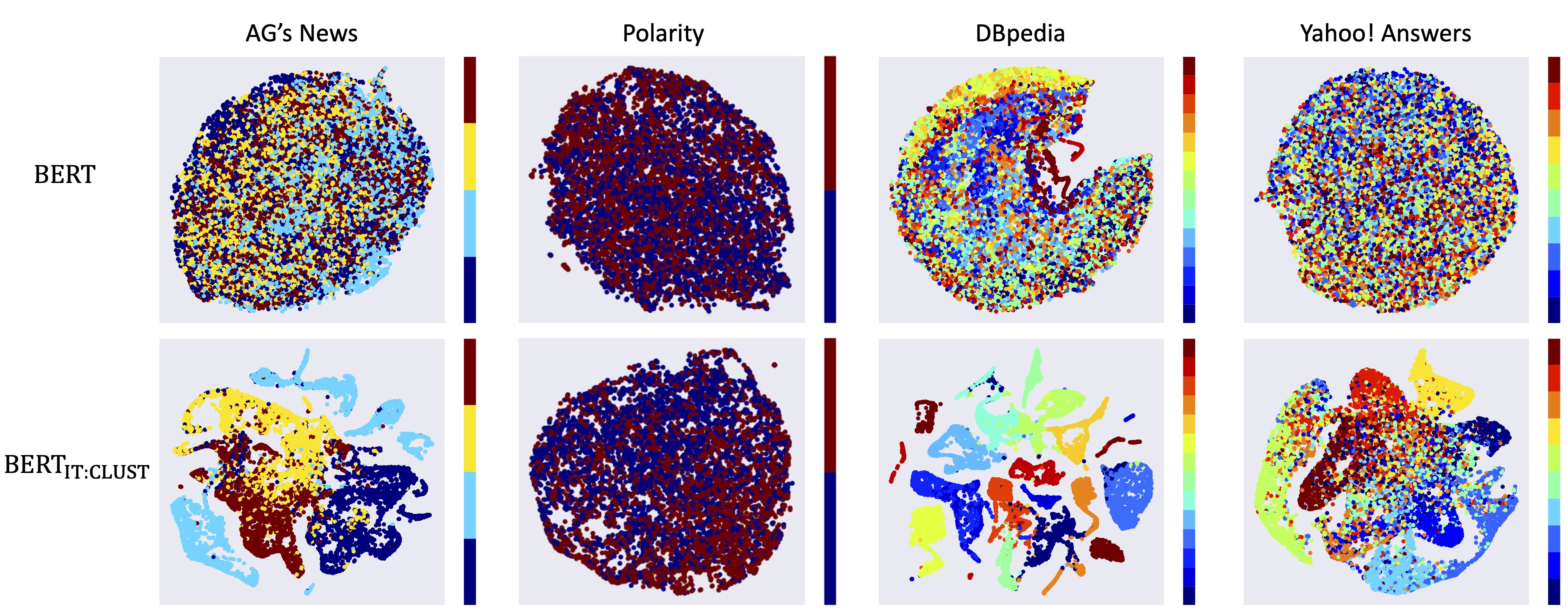}

\caption{t-SNE visualizations of model embeddings over the train set, using BERT (top) vs. \itclust{} (bottom). The colors represent the gold labels for the target task (e.g., four classes in AG's News data set).
}
  \label{fig:tsne}
\end{figure*}

\subsection{Additional Clustering Techniques}\label{ssec:kmeans}

In the literature \citep{sib:02} and on our initial trials, \sib{} showed better clustering performance, and therefore was chosen over other clustering methods. Next, we analyze whether \sib{} is also the best fit for inter-training.

We compare (see App.~\ref{app:more-clustering}) \sib{} over BOW representation to two other clustering configurations; K-means over GloVe representations
and Hartigan's K-means \citep{hartigan} over GloVe. 
For most datasets, inter-training over the results of \sib{} over BOW representations achieved the best results.

\subsection{Comparison to BOW-based methods}
Our inter-training method relies on BOW-based clustering. Since knowledge of the input words is potentially quite powerful for some text classification tasks, we examine the performance of several BOW-based methods. 
We used the same training samples to train multinomial Naive Bayes (NB) and Support Vector Machine (SVM) classifiers, using either Bag of Words (BOW) or GloVe \citep{Pennington2014GloveGV} representations. For GloVe, a text is represented as the average GloVe embeddings of its tokens.
This yielded four reference settings: NB$_{\textrm{BOW}}$, NB$_{\textrm{GloVe}}$, SVM$_{\textrm{BOW}}$ and \glovesvm{}. Overall, all four methods were inferior to \itclust{}, as shown in App.~\ref{app:baselines}. Thus, the success of our method cannot simply be attributed to the information in the BOW representations.

Next, we inspect the contribution of inter-training to BERT's sentence representations.

\subsection{Effect on Sentence Embeddings}\label{ssec:embeddings}

The embeddings after \itclust{} show potential as a better starting point for fine-tuning. Figure \ref{fig:tsne} depicts t-SNE \citep{tSNE} 
2D visualizations of the output embeddings over the full train set of several datasets, 
comparing the [CLS] embeddings before and after inter-training. 

Manifestly, for topical datasets, the \itclust{} embeddings, obtained after inter-training with respect to \sib{} clusters, induce a much clearer separation between the target classes, even though no labeled data was used to obtain this model. Moreover, and perhaps not surprisingly, the apparent visual separation resulting from inter-training is aligned with the performance gain obtained later on in the fine-tuning phase over the target task (as seen, for instance, in the visualizations of Polarity versus DBpedia data). 

In addition to the qualitative results of the visualization, we pursue a more quantitative path. We assess whether examples of the same class are more closely represented after inter-training. Formally, given a set of instances' embeddings $e_1,\ldots, e_n$ and their corresponding class labels $l_1, \ldots, l_n \in \mathcal{L}$ we compute for each class $l\in\mathcal{L}$ a centroid $c_l$ which is the average embedding of this class. 
We then compute the average Euclidean \textit{Embeddings' Distance} (\textit{ED}) from the corresponding centroids:\footnote{Macro average results were similar, we hence report only micro average results. 
Results with Cosine similarity were also similar, hence omitted.} $$ED(l,e)=\mathbb{E}_{i=0}^n \lVert e_i-c_i \rVert_2$$

As a sanity check, we apply a significance test to the ED statistic, confirming that representations of same-class examples are close to each other. Specifically, we apply a permutation test \citep{Fisher1971StatisticalMF}, with $1000$ repetitions, comparing the class labels to random labels.
We find that EDs for both BERT and \itclust{} are significantly different from random ($p<0.001$). 
This implies that both before and after inter-training, same-class representations are close.
Next, we compare the representations before and after inter-training. We find that the randomly permuted EDs of \itclust{} are about 3 times larger than BERT's, despite similar norm values. 
This means that the post inter-training representations are more dispersed. 
Hence, to properly compare, we normalize ED by the average of the permuted EDs: $$NED(l,e)=\frac{ED(l,e)}{\mathbb{E}_{\tau\in S_n}ED(\tau(l),e)}$$ 
Where $\tau\in S_n$ is a permutation out of $S_n$ the set of all permutations.

Comparing the \textit{Normalized Embeddings' Distance} (\textit{NED}) before and after inter-training, we find that in all datasets 
the normalized distance is smaller after inter-training. In other words, \itclust{} brings same-class representations closer in comparison to BERT.

\subsection{Are Clusters Indicative of Target Labels?}\label{ssec:NMI}
\begin{figure}[bht]

\includegraphics[trim={0 0 0 1.45cm},clip, width=.54\textwidth]{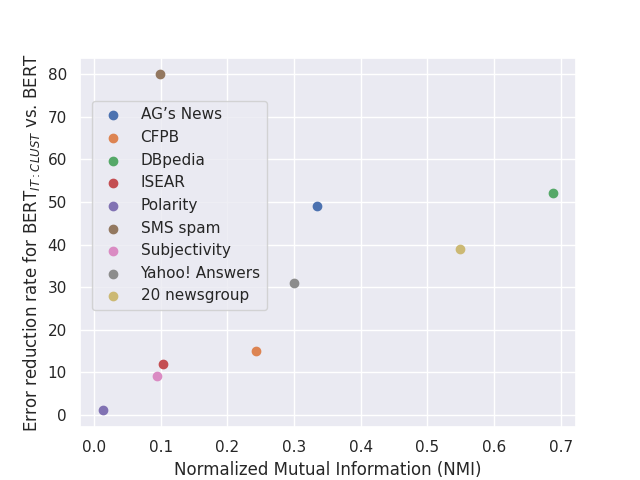}

\caption{Improvement by \itclust{} vs Normalized Mutual Information (NMI) per dataset. x-axis: NMI between the 
cluster and class labels, over the train set. y-axis: The error reduction (percentage) by \itclust{}, when fine-tuning over $64$ samples.}

  \label{fig:cluster-analysis}
  
\end{figure}

A natural explanation for the contribution of inter-training to BERT's performance is that the pseudo-labels, obtained via the clustering partition, are informative with regards to target task labels. To quantify this intuition, in Figure \ref{fig:cluster-analysis} 
we depict the Normalized Mutual Information (NMI) between \sib{} labels and the target task labels, calculated over the entire training set, versus the
gain of using \itclust{}, reflected as the reduction in classification error rate between BERT and \itclust{}, at the extreme case of $64$ fine-tuning samples. 
Evidently, in datasets where the NMI is around zero, \itclust{} does not confer a clear benefit; conversely, where the NMI is relatively high, the performance gains are pronounced as well. Notably, the three datasets with the lowest NMI are those for which inter-training was not beneficial, as discussed in Section~\ref{sec:results}. 

Since the partition obtained via clustering is often informative for the target class labels, we examine whether it can be utilized directly, as opposed to as pseudo-labels for BERT inter-training. To that end, we applied a simple heuristic. Given a labeling budget $x$, we divide it across clusters, relative to their size, while ensuring that at least one instance within 
each of the 50 clusters is labeled. We use the budget per cluster to reveal the labels of a random sample of examples in that cluster, and identify each cluster with its most dominant label. Next, given a new test example, we assign it with the label associated with its nearest cluster. 
Results 
(see App.~\ref{app:baselines})
showed that this rudimentary classifier is generally not on par with \itclust{}, 
yet it can be surprisingly effective where the NMI is high and the labeling budget is small.

\section{Related Work} \label{sec:background}

In our work, we transfer a pretrained model to a new domain with little data.
Transfer learning studies how to transfer models across domains. It suggests methods such as pivoting \citep{Ziser2018PivotBL}, weak supervision \citep{shnarch2018will}, data augmentation \citep{AnabyTavor2020DoNH} and adversarial transfer \citep{Cao2018AdversarialTL}. 

In Computer Vision, pretrained models are often learnt by image clustering  \citep{Caron2018DeepCF}. In NLP, however, clustering was mainly used for non-transfer scenarios. \citet{Ball2019BrownUR} relies on pretrained embeddings to cluster labeled and unlabeled data. Then, they fill the missing labels to augment the training data. 
Clustering itself was improved by combining small amounts of data \citep{torres2019cl, wang2016semi}. 

Pretrained models improved state-of-the-art in many downstream tasks \citep{nogueira2019passage, ein2020corpus} and they are especially needed and useful in low resource 
and limited labeled data settings \citep{Lacroix2019NoisyCF, Wang2020ToPO, Chau2020ParsingWM}. 
There are many suggestions to improve such models, including larger models \citep{Raffel2019ExploringTL}, changes in the pretraining tasks and architecture \citep{Yang2019XLNetGA}, augmenting pretraining \citep{Geva2020InjectingNR}, or improving the transfer itself \citep{Valipour2019UnsupervisedTL,Wang2019ToTO,Sun2019HowTF,Xu2020ImprovingBF}.
Two findings on pretraining support our hypothesis on the intermediate task, namely that classification surpasses MLM.
Some pretraining tasks are better than others \citep{ALBERT, Raffel2019ExploringTL} and
supervised classification as additional pre-training improves performance \citep{Lv2020PretrainingTR, wang-etal-2019-tell,pruksachatkun2020intermediatetask}. 
All these works aim to improve the performance upon transfer, making it more suitable for any new domain. In contrast, we focus on improvement given the domain.

With a transferred model, one can further improve performance with domain-specific information. For example, utilizing metadata \citep{Melamud2019CombiningUP}, training on weakly-supervised data \citep{Raisi2018WeaklySC, meng2020text} or multitasking on related tasks concurrently \citep{Liu2019MultiTaskDN}. Given no domain-specific information, it was suggested to further pretrain on unlabeled data from the domain \citep{Whang2019DomainAT, xu2019bert, Sung2019PreTrainingBO, Rietzler2020AdaptOG, Lee2020BioBERTAP, Gururangan2020DontSP}. This, however, is sometimes unhelpful or even hurts results \citep{Pan2019AnalyzingBW}. 

Transferring a model and retraining with paucity of labels is often termed few-shot learning. Few shot learning is used for many language-related tasks such as named entity recognition \citep{Wang2020AdaptiveSF}, relation classification \citep{Hui2020FewshotRC}, and parsing \citep{schuster2019cross}. There have also been suggestions other than fine-tuning the model. \citet{Koch2015SiameseNN} suggests ranking examples' similarity with Siamese networks. \citet{vinyals2016matching} rely on memory and attention to find neighboring examples and  \citet{snell2017prototypical} search for prototypes to compare to.  \citet{Ravi2017OptimizationAA} don't define in advance how to compare the examples. Instead, they meta-learn how to train the few shot learner. These works addressed the image classification domain, but they supply general methods which are used, improved and adapted on language domains \citep{Geng2019InductionNF, yu2018diverse}.

In conclusion, separate successful practices foreshadow our findings: Clustering drives pre-training on images; supervised classification aids pre-training; and training on unlabeled domain examples is helpful with MLM.
\section{Conclusions} \label{sec:disc}

We presented a simple approach for improving pre-trained models for text classification. Specifically, we show that inter-training BERT over pseudo-labels generated via unsupervised clustering creates a better starting point for the final fine-tuning over the target task. Our analyses suggest that BERT can leverage these pseudo-labels, namely that there exists a beneficial interplay between the proposed inter-training and the later fine-tuning stage. Our results show that this approach yields a significant boost in accuracy, 
mainly over topical data and  
when labeled data is scarce.
Note that the method does require the existence of an unlabeled corpus, in the order of several thousand examples.

We opted here for a practically oriented approach, which we do not claim to be optimal. Rather, the success of this approach suggests various directions for future work. In particular, several theoretical questions arise, such as what else determines the success of the approach in a given dataset; 
understanding the potential synergistic effect of BOW-based clustering for inter-training; 
could more suitable partitions be acquired by 
exploiting additional embedding space and/or more clustering techniques
; co-training \citep{Co-training} methods, 
and more. 

On the practical side, while in this work we fixed the inter-training to be over $50$ clusters and for a single epoch, future work can improve performance by tuning such hyper-parameters.
In addition, one may consider using the labeled data available for fine-tuning as anchors for the intermediate clustering step, which we have not explored here. 

Another point to consider is the nature of the inter-training task. Here, we examined a multi-class setup where BERT is trained to predict one out of $n_c$ cluster labels. Alternatively, one may consider a binary inter-training task, where BERT is trained to determine whether two samples are drawn from the same cluster or not. 

Finally, the focus of the present work was on improving BERT performance for text classification. In principle, inter-training BERT over clustering results may be valuable for additional downstream target tasks, that are similar in spirit to standard text classification. Examples include Key-Point Analysis \citep{KPA} and Textual Entailment \citep{TE}.
The potential value of our approach in such cases is left for future work.

\section*{Acknowledgements}
We thank Assaf Toledo for providing helpful advice on the clustering implementations.

\clearpage
\bibliography{custom}
\bibliographystyle{acl_natbib}

\appendix
\section{Datasets} \label{app:datasets}

Links for downloading the datasets:
\begin{description}
    \item [Polarity:] \url{http://www.cs.cornell.edu/people/pabo/movie-review-data/}.
    
    \item [Subjectivity:] \url{http://www.cs.cornell.edu/people/pabo/movie-review-data/}.
    
    \item [CFPB:] \url{https://www.consumerfinance.gov/data-research/consumer-complaints/}.

    \item [20 newsgroups:]
    \url{http://qwone.com/~jason/20Newsgroups/}\\
    We used the version provided by scikit: \url{https://scikit-learn.org/0.15/datasets/twenty_newsgroups.html}. 
    
    \item [AG’s News, DBpedia and Yahoo! answers:]
    We used the version from: \url{https://pathmind.com/wiki/open-datasets} (look for the link \textit{Text Classification Datasets}).

    \item [SMS spam:] \url{http://www.dt.fee.unicamp.br/~tiago/smsspamcollection/}
    
    \item [ISEAR:]  \url{https://www.unige.ch/cisa/research/materials-and-online-research/research-material/}.
    
\end{description}

\section{Additional reference methods}\label{app:baselines}
\begin{figure*}
\center

\makebox[\textwidth][c]{
  \includegraphics[width=0.33\textwidth]{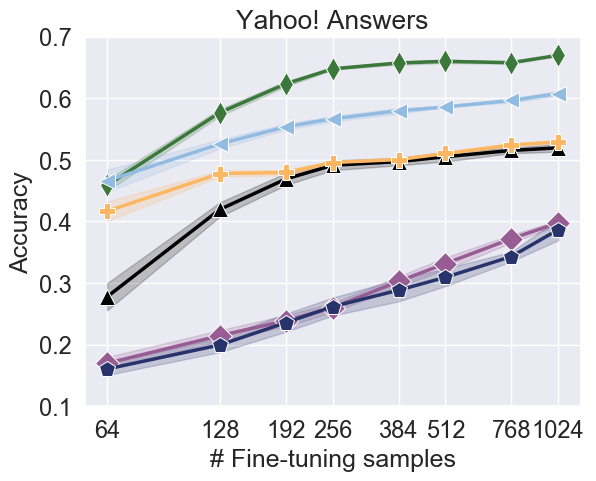}
  \includegraphics[width=0.33\textwidth]{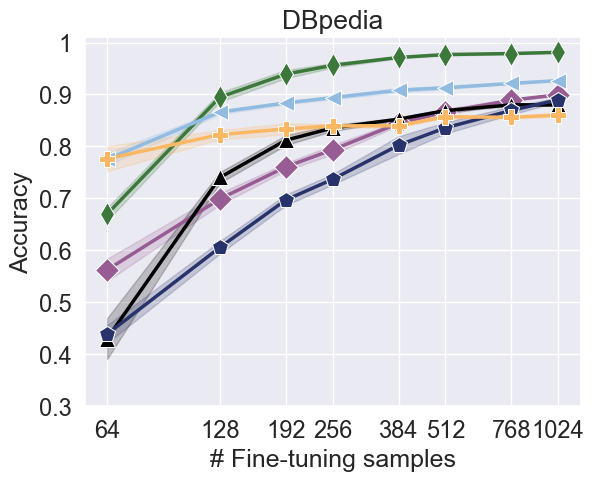}
    \includegraphics[width=0.33\textwidth]{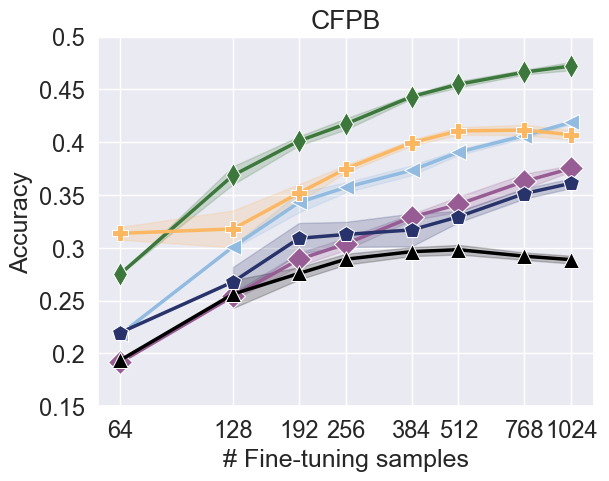}
}
\makebox[\textwidth][c]{  
  \includegraphics[width=0.33\textwidth]{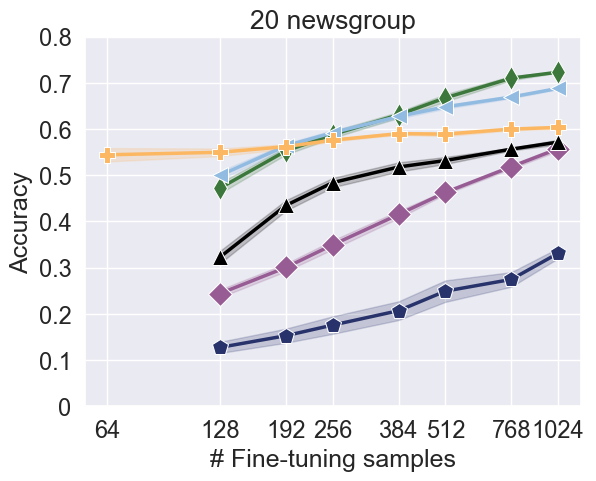}
  \includegraphics[width=0.33\textwidth]{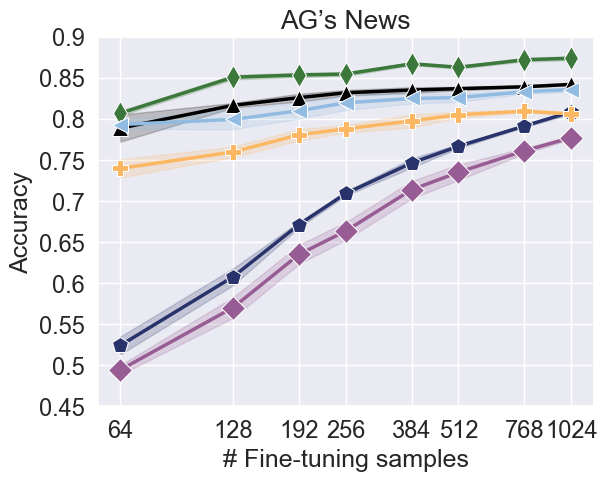}
  \includegraphics[width=0.33\textwidth]{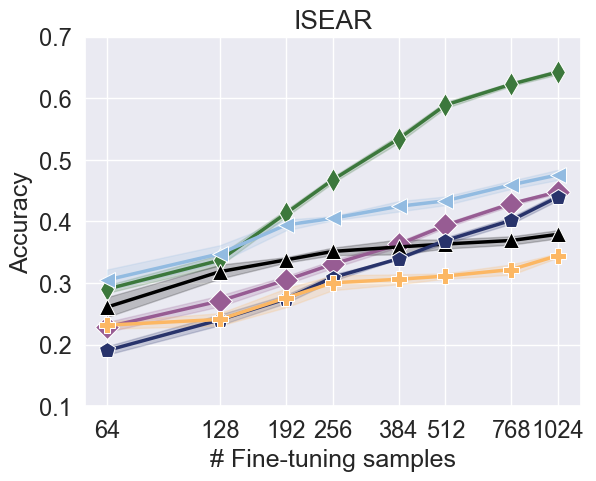}
}
\makebox[\textwidth][c]{
    \includegraphics[width=0.33\textwidth]{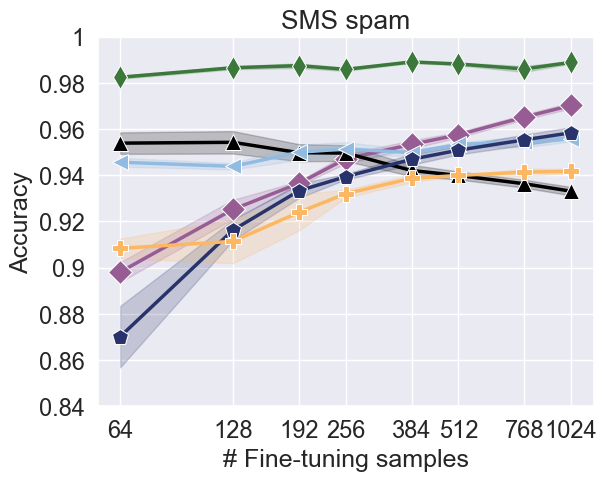}
    \includegraphics[width=0.33\textwidth]{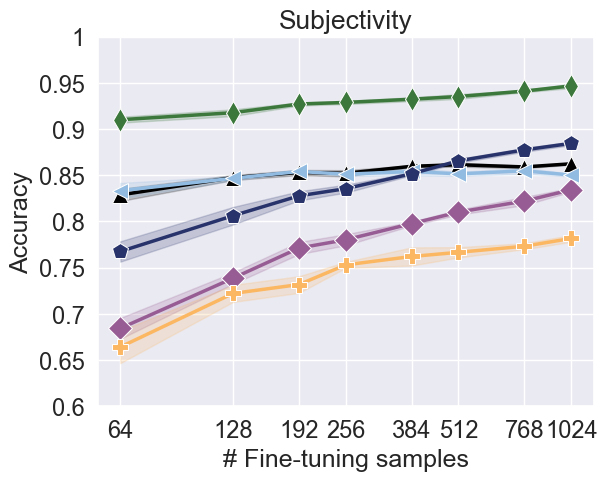} 
    \includegraphics[width=0.33\textwidth]{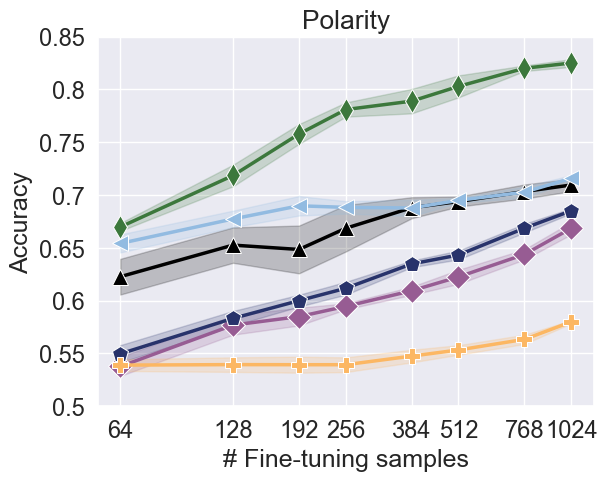}
}
\makebox[\textwidth][c]{
    \includegraphics[width=\textwidth]{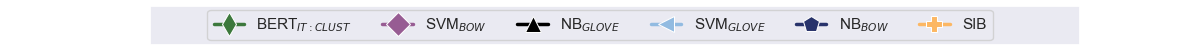}
}

\caption{Comparing BOW methods and the \itclust{} setting. Each point is the average of five repetitions ($\pm$ SEM). X axis denotes the budget for training in log scale, and Y accuracy of each model. 
}

  \label{app-fig:baselines}
  
\end{figure*}

The results of NB$_{\textrm{BoW}}$, NB$_{\textrm{GloVe}}$, SVM$_{\textrm{BoW}}$ and \glovesvm{} are shown in Figure \ref{app-fig:baselines}.

\paragraph{\sib{}-based classifier}
As mentioned in \S\ref{ssec:NMI}, we experimented with building a rudimentary classifier that utilizes only the \sib{} clustering results and the labeling budget. 
Results for this setting are depicted in Fig.~\ref{app-fig:baselines} in orange. Comparing these results to the BERT-based approaches reveals that clustering alone is not sufficient.

\section{Additional clustering techniques}\label{app:more-clustering}
\begin{figure*}
\center
\makebox[\textwidth][c]{
  \includegraphics[width=0.33\textwidth]{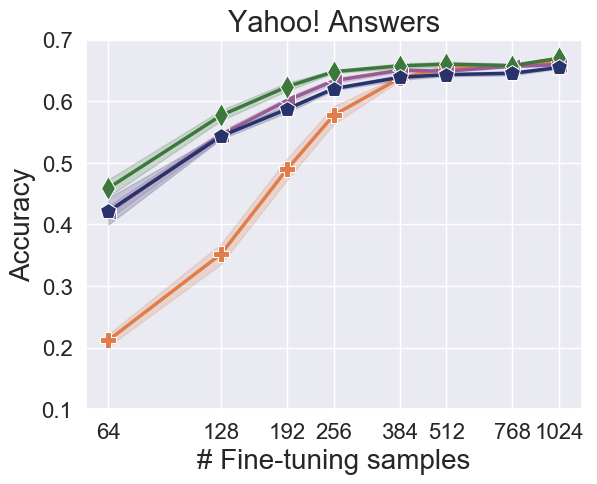}
  \includegraphics[width=0.33\textwidth]{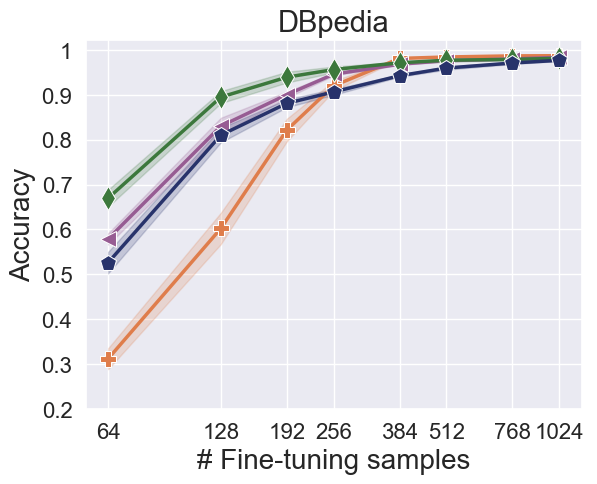}
  \includegraphics[width=0.33\textwidth]{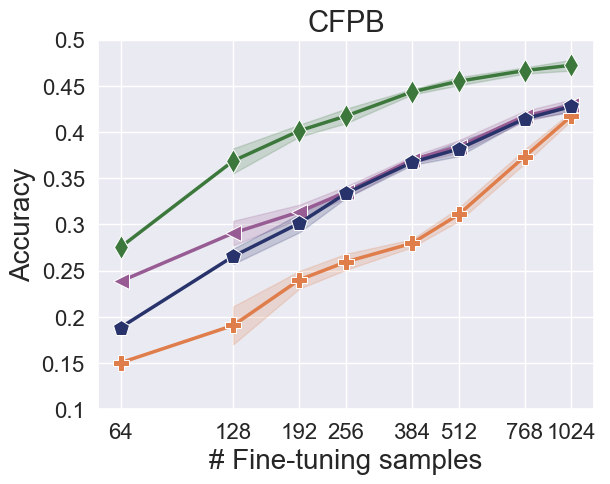}
}
\makebox[\textwidth][c]{
  \includegraphics[width=0.33\textwidth]{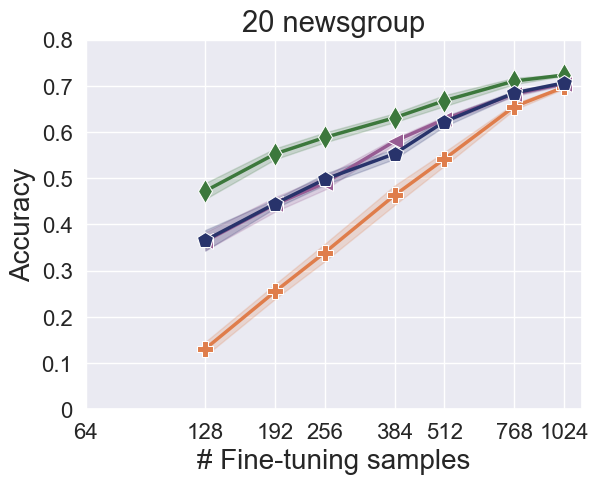}
    \includegraphics[width=0.33\textwidth]{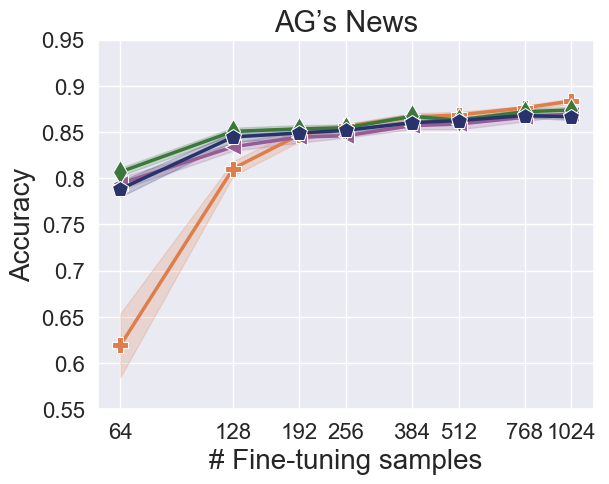}
    \includegraphics[width=0.33\textwidth]{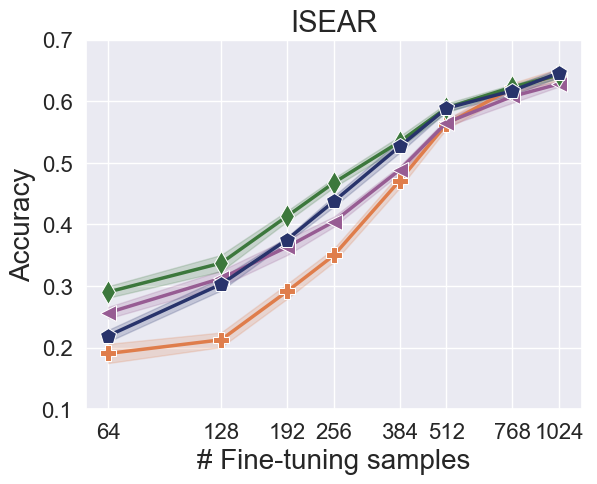}
}
\makebox[\textwidth][c]{
    \includegraphics[width=0.33\textwidth]{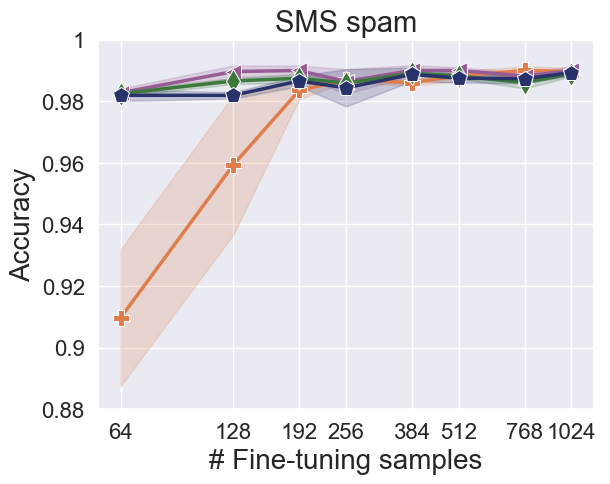}
    \includegraphics[width=0.33\textwidth]{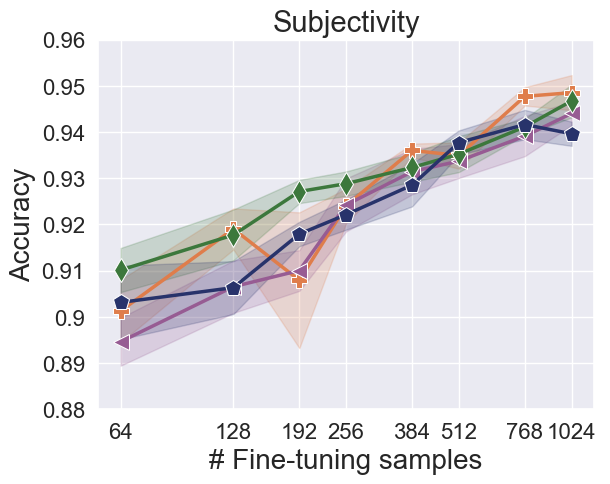}
    \includegraphics[width=0.33\textwidth]{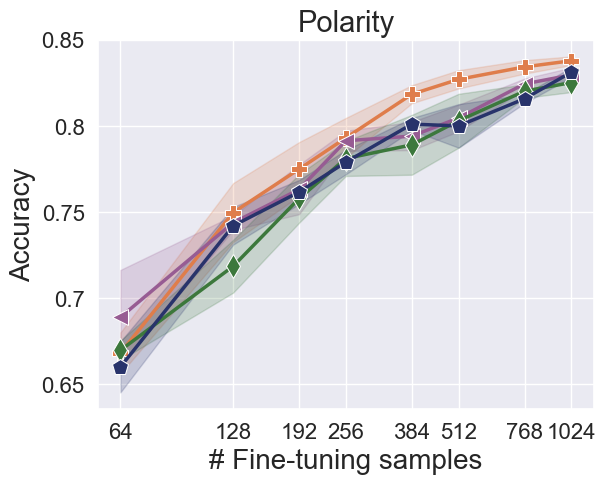}
}
\makebox[\textwidth][c]{
    \includegraphics[width=\textwidth]{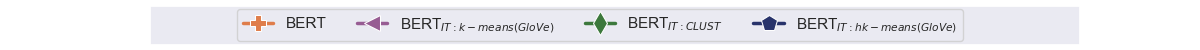}
}

\caption{Comparison of clustering configurations for the intermediate task (hk-means stands for Hartigan's K-means). The results with no inter-training (BERT) are also presented for comparison. 
Each point is the average of five repetitions ($\pm$ SEM).  X axis denotes the number of labeling instances used for fine-tuning (in log scale).}

  \label{fig:more-clustering}
  
\end{figure*}

Fig.~\ref{fig:more-clustering} depicts the comparison of the \sib{} over BOW representation, denoted \itclust{}, to two other configurations for the clustering intermediate task: K-means over GloVe representations and Hartigan's K-means \citep{hartigan} over GloVe. The GloVe representation for each text is an average of GloVe representations for the individual tokens.
The comparison reveals that in most cases \sib{} over BOW outperforms the other clustering configurations.

\FloatBarrier

\end{document}